\documentclass{article}

\PassOptionsToPackage{numbers, compress}{natbib}

\usepackage[preprint]{nips_2018}

\usepackage[utf8]{inputenc} 
\usepackage[T1]{fontenc}    
\usepackage{hyperref}       
\usepackage{url}            
\usepackage{booktabs}       
\usepackage{amsfonts}       
\usepackage{nicefrac}       
\usepackage{microtype}      

\usepackage{mathtools}
\usepackage{multirow}
\usepackage{amsthm}
\usepackage{units}
\usepackage{diagbox}

\newcommand{\eg}{e.\,g., }
\newcommand{\ie}{i.\,e. }

\newtheorem{definition}{Definition}

\title{Message Passing Graph Kernels}

\author{
  Giannis Nikolentzos \\
  \'Ecole Polytechnique\\
  \texttt{nikolentzos@lix.polytechnique.fr} \\
  \And
  Michalis Vazirgiannis \\
  \'Ecole Polytechnique\\
  \texttt{mvazirg@lix.polytechnique.fr}
}

\begin{document}

\maketitle

\begin{abstract}
Graph kernels have recently emerged as a promising approach for tackling the graph similarity and learning tasks at the same time.
In this paper, we propose a general framework for designing graph kernels.
The proposed framework capitalizes on the well-known message passing scheme on graphs.
The kernels derived from the framework consist of two components.
The first component is a kernel between vertices, while the second component is a kernel between graphs. 
The main idea behind the proposed framework is that the representations of the vertices are implicitly updated using an iterative procedure.
Then, these representations serve as the building blocks of a kernel that compares pairs of graphs.
We derive four instances of the proposed framework, and show through extensive experiments that these instances are competitive with state-of-the-art methods in various tasks.
\end{abstract}

\section{Introduction}

Graph-structured data arises naturally in many domains ranging from bioinformatics and social networks to
cybersecurity.
A key issue in many applications is to perform machine learning tasks on this type of data.
In the past years, the problem of graph classification has found applications in several fields such as in chemoinformatics \cite{ralaivola2005graph}, in malware detection \cite{gascon2013structural} and in text categorization \cite{nikolentzos2017shortest}.
For instance, in chemoinformatics, molecules are commonly represented as graphs where vertices correspond to atoms and edges to chemical bonds between them.
The task is then to predict the class label of each graph (\eg its anti-cancer activity).

Graph kernels have recently evolved into the dominant approach for learning on graph-structured data.
A graph kernel is a positive semidefinite function defined on the space of graphs $\mathcal{G}$.
This function corresponds to an inner product in some Hilbert space.
Given a kernel $k$, there exists a map $\phi : \mathcal{G} \rightarrow \mathcal{H}$ into a Hilbert space $\mathcal{H}$ such that $k(G_1,G_2) = \langle \phi(G_1), \phi(G_2) \rangle$ for all $G_1,G_2 \in \mathcal{G}$.
One of the major advantages of graph kernels is that they allow kernel methods such as the Support Vector Machines (SVM) to work directly on graphs.
Research in graph kernels has achieved a remarkable progress in the past years.
However, graph kernels have been applied mainly to graphs that are either unlabeled or contain discrete node labels.
For such kind of graphs, there exist several highly scalable graph kernels which can handle graphs with thousands of vertices (\eg the Weisfeiler-Lehman subtree kernel \cite{shervashidze2011weisfeiler}).
However, graphs that emerge from several real settings typically contain multi-dimensional vertex attributes (a.k.a. features).
Such types of graphs appear in computer vision \citep{harchaoui2007image} and in bioinformatics \citep{borgwardt2005protein}, among others.
For instance, in computer vision, attributes may represent the RGB values of colors, while in bioinformatics, they may represent physical properties of protein secondary structure elements.
When continuous node labels are available, taking them into account usually leads to significant performance improvements.
Designing graph kernels for such types of graphs is however a much less well studied problem which started to gain some attention recently \citep{feragen2013scalable,neumann2016propagation,orsini2015graph,kriege2012subgraph,morris2016faster}.
Unfortunately, most of the proposed apporaches do not scale even to relatively small datasets consisting of graphs with tens of vertices.
An open challenge is thus to develop scalable kernels for graphs with continuous-valued multi-dimensional vertex attributes.


Very recently, research in machine learning on graphs shifted towards neural network architectures.
Several neural network models have been generalized to work on graph-structured data \cite{niepert2016learning,lee2017deep,zhang2018end,li2015gated,kipf2016semi,battaglia2016interaction}.
In contrast to graph kernels, these networks can efficiently take into account continuous node attributes.
Several of these networks fall under the general class of message passing neural networks \cite{gilmer2017neural}.
The main idea behind these methods is that each vertex receives messages from its neighbours and utilizes these messages to update its representation.
This is a well-established idea which has been widely applied in graph mining.
In fact, this even constitutes the key underlying principle of some graph kernels (\eg Weisfeiler-Lehman kernel \cite{shervashidze2011weisfeiler}, propagation kernel \cite{neumann2016propagation}).

In this paper, we present a new framework for designing graph kernels, called Message Passing graph kernels (MPGK).
The proposed framework consists of two components: ($1$) a kernel that compares vertices, and ($2$) a kernel that compares graphs.
The first component is computed for a number of iterations.
At each iteration, the representation of each vertex is updated implicitly based on its own representation and the representations of its neighbors.
Although this idea has been already explored in the past, we provide a radically different formulation.
The proposed framework is very general, and the whole computation is performed in the kernel space.
In contrast to previous approaches, the proposed framework is capable of handling any type of graphs, while it can more effectively capture similarities between rooted subgraphs.
In contrast to the message passing neural network architectures, the proposed framework uses more sophisticated functions for updating vertex rerpesentations.
One of the drawbacks of the proposed farmework is that its instances suffer from high computational complexity.
Since efficient computation is central for the applicability of the farmework to read-world datasets, we propose an approximation method which reduces the number of evaluations of the kernel between vertices which allows the framework to scale to large datasets.

The rest of this paper is organized as follows.
Section~\ref{sec:preliminaries} introduces some preliminary concepts of message passing approaches.
Section~\ref{sec:contribution} presents the proposed framework for designing message passing graph kernels.
Section~\ref{sec:experiments} evaluates the proposed framework in several tasks and compares it with existing methods.
Finally, Section~\ref{sec:conclusion} concludes.

\section{Preliminaries}\label{sec:preliminaries}
Several approaches that deal with the problem of learning on graph-structured data generate fixed dimensional vector representations for small subgraphs extracted from the input graphs.
For instance, many recent neural networks for graphs collect each node's $k$-hop neighborhood and then generate representations for them.
Since there is no correspondence between the neighbors of different vertices, someone either has to impose an order on the neighboring vertices or to employ a permutation invariant function.
To impose an order on the vertices, it is common to apply labeling procedures (\eg degree, eigenvector centrality, etc.) \cite{niepert2016learning}.
On the other hand, message passing neural networks summarize the neighborhood of a vertex using permutation invariant functions \cite{gilmer2017neural}.
Given a set $\mathcal{X}$, such functions take as input the power set $2^\mathcal{X}$, and produce a function that is independent of the ordering of the elements of the input.
More formally,
\begin{definition}
  A function $f : 2^\mathcal{X} \rightarrow \mathcal{Y}$ acting on sets is permutation invariant to the order of the objects in the set if for any permutation $\pi$ it holds that $f(\{x_1,\ldots,x_n\}) = f(\{x_{\pi(1)},\ldots,x_{\pi(n)}\})$.
\end{definition}

Message passing neural networks do not operate on fixed dimensional vectors, but they take into account the whole set of neighbors of each vertex.
Hence, to aggregate neighborhood information, they employ functions defined on sets that are invariant to permutations.
The majority of these architectures achieve invariance by simply summing the messages coming from each neighbor.
Let $G=(V,E)$ be a graph.
Let also $\mathcal{N}(v)$ be the set of neighbors of vertex $v \in V$.
We denote as $\mathbf{x}_v^t \in \mathbb{R}^{d}$ the representation of vertex $v$ at layer $t$.
Then, most message passing architectures update the representation of each vertex based on the representations of its neighbors.
More specifically, during the message passing phase, hidden states $\mathbf{x}_v^t$ at each vertex in the graph are updated based on messages
$\mathbf{m}_v^{t+1}$ according to:
\begin{equation}
  \begin{split}
    \mathbf{m}_v^{t+1} &= \sum_{u \in \mathcal{N}(v)} M_t(\mathbf{x}_v^t, \mathbf{x}_u^t) \\
    \mathbf{x}_v^{t+1} &= U_t(\mathbf{x}_v^t, \mathbf{m}_v^{t+1})
  \end{split}
\end{equation}
The above update strategy illustrates the major weakness of such neural networks.
Taking the sum of the messages sent from each neighbor is clearly a permutation invariant function since the response of the function is ``indifferent'' to the ordering of the elements.
However, due to its simplistic nature, this function poses a serious limitation that restricts the representation power of message passing neural networks.
Hence, it is clear that more sophisticated approaches are required to learn meaningful vertex representations and as a consequence, meaningful graph representations.



In contrast to the message passing neural networks, some graph kernels are capable of learning more expressive representations for the neighborhood of each vertex.
However, some of them operate only on graphs with discrete node labels \cite{shervashidze2011weisfeiler}, while others employ very expensive procedures such as the Bhattacharyya kernel \cite{kondor2003kernel} to compare the neighborhood graphs \cite{kondor2016multiscale}.

\section{Message Passing Graph Kernels}\label{sec:contribution}
In this Section, we introduce a message passing framework for comparing graphs.
Due to the sophisticated permutation invariant kernel functions it employs, the framework is more expressive than neural message passing architectures.
We propose an iterative procedure that propagates vertex representations.
The framework assumes that each vertex is assigned an initial representation (either a discrete label or continuous attributes). 
In the case of unlabeled graphs, the representation of each vertex can be initialized using local vertex features.
Such features include for instance, the degree of the vertex, the number of triangles in which the vertex participates, etc.

The proposed framework consists of two components.
The first component is a kernel between vertices and the second component a kernel between graphs.
Note that the first component allows someone to perform machine learning tasks at the node level, while combined with the second component, it allows someone to perform machine learning tasks at the graph level.
Let $k_v$ be a kernel between vertices and $k_{\mathcal{N}(v)}$ a kernel between neighborhoods.
Then, the proposed framework computes iteratively a kernel $k_v^t$ between each pair of vertices, where $t$ denotes the timestep.
Specifically, the kernel values between the vertices are updated following the recurrence shown below:
\begin{equation}
  \label{eq:vertex_update}
    k_v^{t+1}(v_1, v_2) = \alpha \; k_v^t(v_1, v_2) + \beta \; k_{\mathcal{N}(v)} \big(\mathcal{N}(v_1), \mathcal{N}(v_2) \big)
\end{equation}
where $\alpha$ and $\beta$ are nonnegative constants.
Clearly, $k_v^{t+1}$ is a positive semidefinite function defined on the space of vertices given that $k_v^t$ and $k_{\mathcal{N}(v)}$ are also positive semidefinite kernels.
It is interesting to note that the above procedure implicitly updates the representations of the considered vertices.
At the first iteration, a kernel function that compares the labels/attributes of the vertices is employed and then, all the computations are performed in the kernel space.
After computing the kernel between each pair of vertices for $T$ iterations, we can compute a kernel between graphs as follows:
\begin{equation}
  \label{eq:graph_update}
    k_G(G_1, G_2) = k_V(V_1, V_2)
\end{equation}
where $k_V$ is a kernel between sets of vertices.
Note that both $k_V$ and $k_{\mathcal{N}(v)}$ are functions defined on sets of vertices, and hence, they are required to satisfy the constraint of permutation invariance.
To compute the above two kernels, we employ two well-known design paradigms for developing kernels: ($1$) the R-convolution framework \cite{haussler1999convolution} and ($2$) the theory of valid optimal assignment kernels \cite{kriege2016valid}.
Given two sets of vertices $V_1$ and $V_2$, we propose the following R-convolution kernel:
\begin{equation}
    k_V(V_1, V_2) = \sum_{v_1 \in V_1} \sum_{v_2 \in V_2} k_v(v_1, v_2)
\end{equation}
and the following assignment kernel:
\begin{equation}
    k_V(V_1, V_2) = \max_{B \in \mathfrak{B}(V_1, V_2)} \sum_{(v_1, v_2) \in B} k_s(v_1, v_2)
\end{equation}
where $\mathfrak{B}(V_1,V_2)$ denotes the set of all bijections between the two sets of vertices $V_1,V_2$ (for simplicity we have assumed that the size of both sets is the same) and $k_s$ is a strong kernel defined on vertices (see \cite{kriege2016valid} for more details).
Both kernels defined above are permutation invariant and are therefore eligible for comparing the vertices of two graphs and/or the neighbors of two vertices.

Based on the above two formulations, the update scheme defined in Equation~\ref{eq:vertex_update} becomes:
\begin{equation}
  \label{eq:vertex_r_conv}
    k_v^{t+1}(v_1, v_2) = \alpha \; k_v^t(v_1, v_2) + \beta \; \sum_{u_1 \in \mathcal{N}(v_1)} \sum_{u_2 \in \mathcal{N}(v_2)} k_v^t(u_1, u_2)
\end{equation}
and
\begin{equation}
  \label{eq:vertex_assign}
    k_v^{t+1}(v_1, v_2) = \alpha \; k_v^t(v_1, v_2) + \beta \; \max_{B \in \mathfrak{B} \big(\mathcal{N}(v_1),\mathcal{N}(v_2) \big)} \sum_{(u_1, u_2) \in B} k_s^t(u_1, u_2)
\end{equation}
respectively.
To compare a pair of graphs, we use the same permuation invariant kernel functions, and Equation~\ref{eq:graph_update} becomes:
\begin{equation}
  \label{eq:graph_r_conv}
    k_G(G_1, G_2) = \sum_{v_1 \in V_1} \sum_{v_2 \in V_2} k_v^T(v_1, v_2)
\end{equation}
and
\begin{equation}
  \label{eq:graph_assign}
  k_G(G_1, G_2) = \max_{B \in \mathfrak{B}(V_1, V_2)} \sum_{(v_1, v_2) \in B} k_s^T(v_1, v_2)
\end{equation}
where $\mathfrak{B}(V_1, V_2)$ denotes the set of all bijections between the sets of vertices of $G_1$ and $G_2$.
Again, we have assumed that the size of the two graphs is the same.

By combining Equations~\ref{eq:vertex_r_conv},~\ref{eq:vertex_assign} with Equations~\ref{eq:graph_r_conv},~\ref{eq:graph_assign} we derive four variants of the proposed framework where neighbor vertices and graph vertices are compared using either an R-convolution or an assignment kernel.
We denote these variants by the abbreviations MPGK RR, MPGK RA, MPGK AR, and MPGK AA; here the letter R stands for the R-convolution kernel and the letter A for the assignment kernel.
The first letter indicates the employed kernel between neighbor vertices and the second letter the employed kernel between graph vertices.

As regards kernel $k_v^0$, for graphs with discrete node labels, we use a delta kernel, while for graphs with continuous node attributes, we use a linear kernel between the nodes' attributes.
To compute the assignment kernel between two sets of vertices (either the sets of neighbors of two vertices or the sets of vertices of two graphs), we capitalize on the methodology of valid assignment kernels \cite{kriege2016valid}. 
Therefore, we define a hierarchy $H=(T,w)$ which induces a strong kernel $k_s$ and this kernel ensures that the emerging assignment function is positive semidefinite.
To build the hierarchy, we resort to clustering.
Specifically, to create the tree $T$, we perform kernel $k$-means using the kernel values between the vertices (those of the previous time step when comparing neighborhoods and those of the last time step when comparing graphs. For $k_s^0$, since there is no previous time step, we first compute the kernel $k_v^0$ between vertices as defined above).
The value of the weighting function $w$ is determined based on the $\omega$ function which is defined as follows: the $\omega$ value of the root $r$, is set equal to $1$.
The $\omega$ value of any other vertex $v \in V(T) \backslash \{r\}$ is set equal to $\omega(v) = \nicefrac{(sp(v,r)-1)}{sp(v,r)}$ where $sp(v,r)$ is the length of the shortest path between the root $r$ and vertex $v$ and $V(T)$ is the set of vertices of the tree $T$.
Hence, as expected, $\omega$ weights more vertices appearing lower in the hierarchy than those appearing higher.

\subsection{Link to Weisfeiler-Lehman Subtree Kernel}
The proposed kernel is related to the Weisfeiler-Lehman subtree kernel \cite{shervashidze2011weisfeiler}, a state-of-the-art kernel for graphs with discrete node labels.
In fact, the Weisfeiler-Lehman subtree kernel can be seen as an instance of the proposed framework.
The Weisfeiler-Lehman subtree kernel uses a combination of message passing and hashing to update the labels of the vertices.
The kernel between two vertices at each time step is equal to the sum of the kernel value of the previous time step and the output of a delta kernel between their labels at that time step.
The kernel between two graphs is the R-convolution kernel defined in Equation~\ref{eq:graph_r_conv}.
In the case of the Weisfeiler-Lehman subtree kernel, the notion of structural equivalence is very rigid since it is defined as a binary property (\ie due to the delta function).
Perturbing the edges by a small amount leads to completely different vertex labels.
Even if two vertices have very similar neighborhoods, it is very likely that they will be assigned different labels after some iterations, and will be thus considered different from each other.
Conversely, in the case of the proposed framework, structural equivalence is not defined as a binary property and the variants we derive are capable of identifying how similar to each other the neighborhoods of two vertices are.

\subsection{Low Rank Approximation}
Given two graphs $G_1$ and $G_2$, computing $k_G^{t+1}(G_1, G_2)$ requires computing $k_v^{t+1}(v_1, v_2)$ between all $\binom{n_1 + n_2}{2}$ pairs of vertices of the two graphs.
Given a dataset that contains $N$ graphs each consisting of $n$ vertices, this translates to computing $k_v^{t+1}(v_1, v_2)$ between all $\binom{nN}{2}$ pairs of vertices.
Furthermore, the complexity of the R-convolution kernel that compares all pairs of neighbors of two vertices is $\mathcal{O}(d_{ave}^2)$ where $d_{ave}$ is the average degree of the vertices of the input graphs.
In the worst scenario (each graph is complete), the average degree of the vertices is $n$, and therefore, the complexity of the R-convolution kernel described above becomes $\mathcal{O}(n^2)$.
Clearly, for large datasets and/or datasets that contain large graphs, evaluating all these kernels between vertices is infeasible.
The complexity of the assignment kernel is lower than that of the R-convolution kernel.
Provided we have already generated the hierarchy inducing kernel $k_s$, the assignment kernel can be computed in time $\mathcal{O}(d_{ave})$.
However, even in that case, the computation of the kernel matrix may still be costly when dealing with large datasets.
In adition, storing the kernel matrix between the $nN$ vertices (\ie an $nN \times nN$-dimensional matrix) may turn out to be infeasible.
To account for that, we resort to approximation algorithms.
Since our goal is to approximate the kernel matrix between the $nN$ vertices, we employ the popular Nystr{\"o}m method \cite{williams2001using}.
It is important to note that the employed method does not require the graphs of the test set to be known during training, but they can be projected to the low-dimensional space at test time.
To compute the kernel value between two graphs, using the R-convolution kernel requires $\mathcal{O}(n^2)$ time, while the assignment kernel can be computed in $\mathcal{O}(n)$ time given a hierarchy inducing the $k_s$ kernel.
The combination of these two factors makes computing the entire stack of kernels feasible.

\section{Experimental Evaluation}\label{sec:experiments}
In this Section, we empirically evaluate the proposed framework on several tasks, and we compare it to state-of-the-art methods.

\subsection{Node Embedding}
We first demonstrate the effectiveness of the proposed framework in learning meaningful vertex representations.
In contrast to many recent methods, the proposed framework can accurately capture the structural identity of vertices.
Conversely, recent approaches for learning node representations such as DeepWalk \cite{perozzi2014deepwalk} and LINE \cite{tang2015line} may fail to encode structural similarity since they mainly take into account the proximity of the vertices in the graph to generate node embeddings.
Therefore, two vertices that are ``far'' from each other in the graph (\eg they belong to different connected components) will also be far from each other in the embedding space, independent of their local structure.
Hence, most recent methods will fail to generate rerpesentations that capture structural equivalence.
One notable exception is struc2vec \cite{ribeiro2017struc2vec} which compares the ordered degree sequences of the nodes' $k$-hop neighborhoods.

\begin{figure}[t]
  \centering
  \includegraphics[trim = 20mm 25mm 30mm 50mm,width=.5\linewidth]{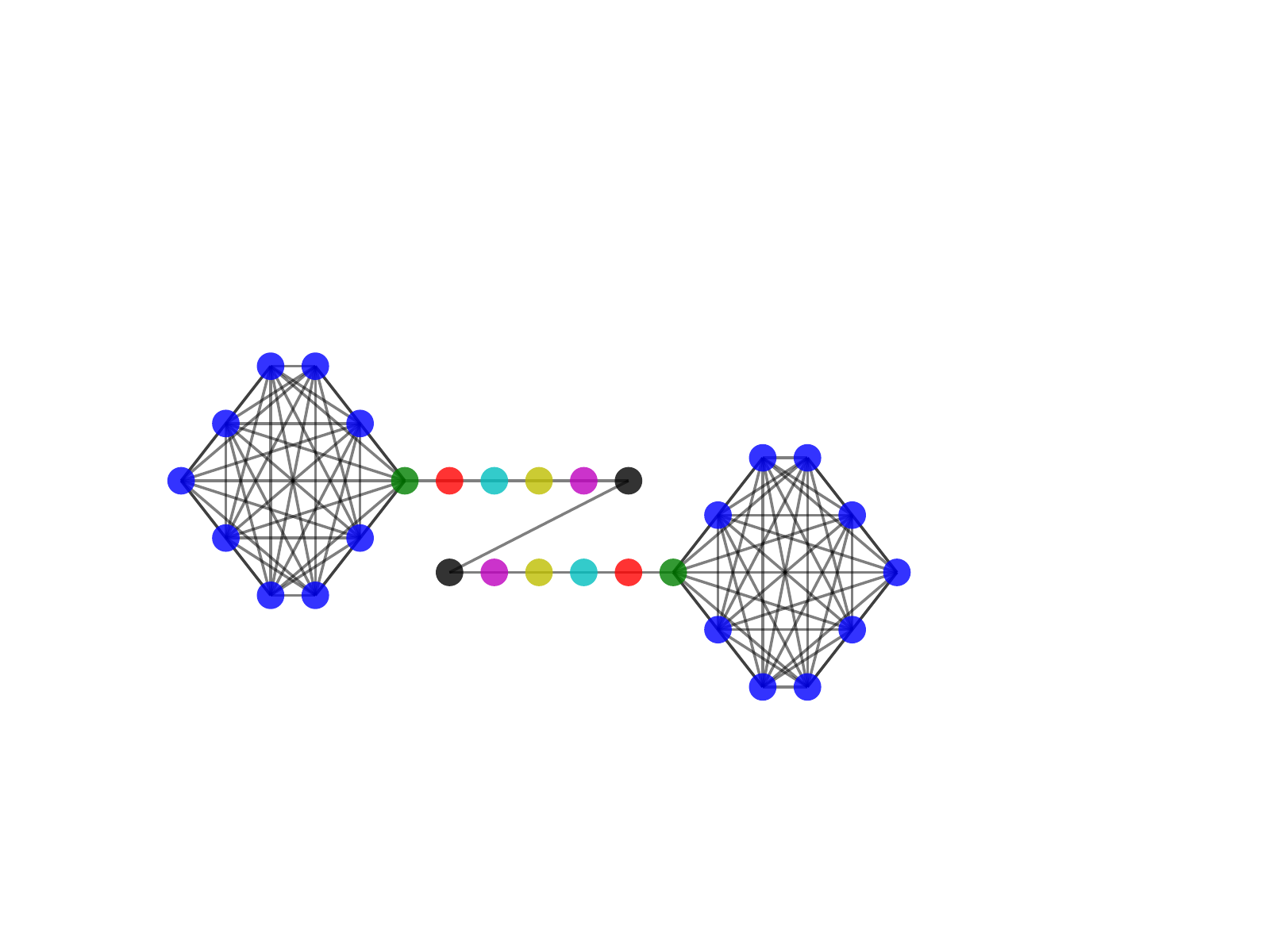}\\
  \begin{minipage}{.5\textwidth}
    \centering
    \texttt{struc2vec}\\
    \includegraphics[width=.8\linewidth]{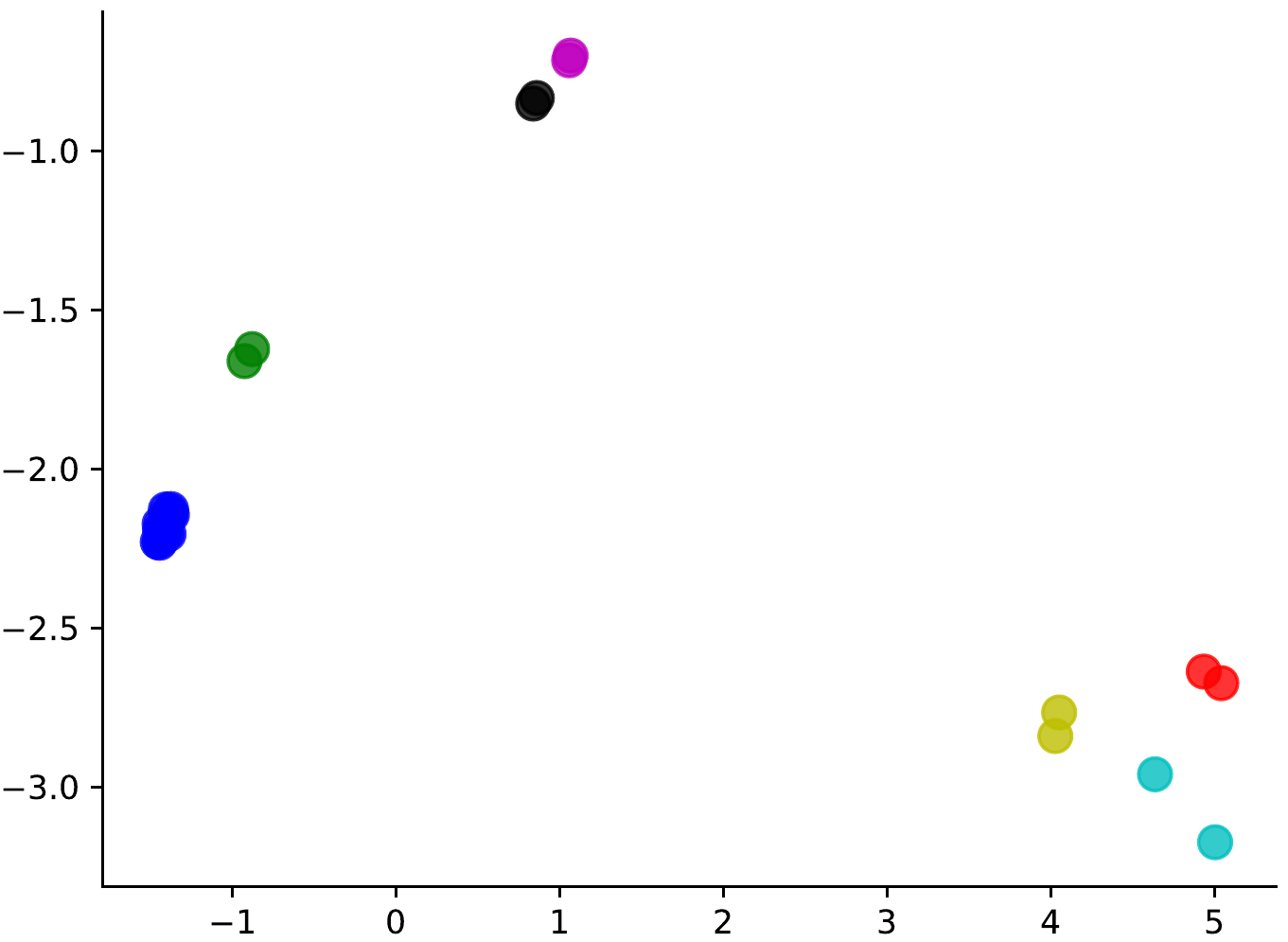}
  \end{minipage}%
  \begin{minipage}{.5\textwidth}
    \centering
    \texttt{MPGK}\\
    \includegraphics[width=.8\linewidth]{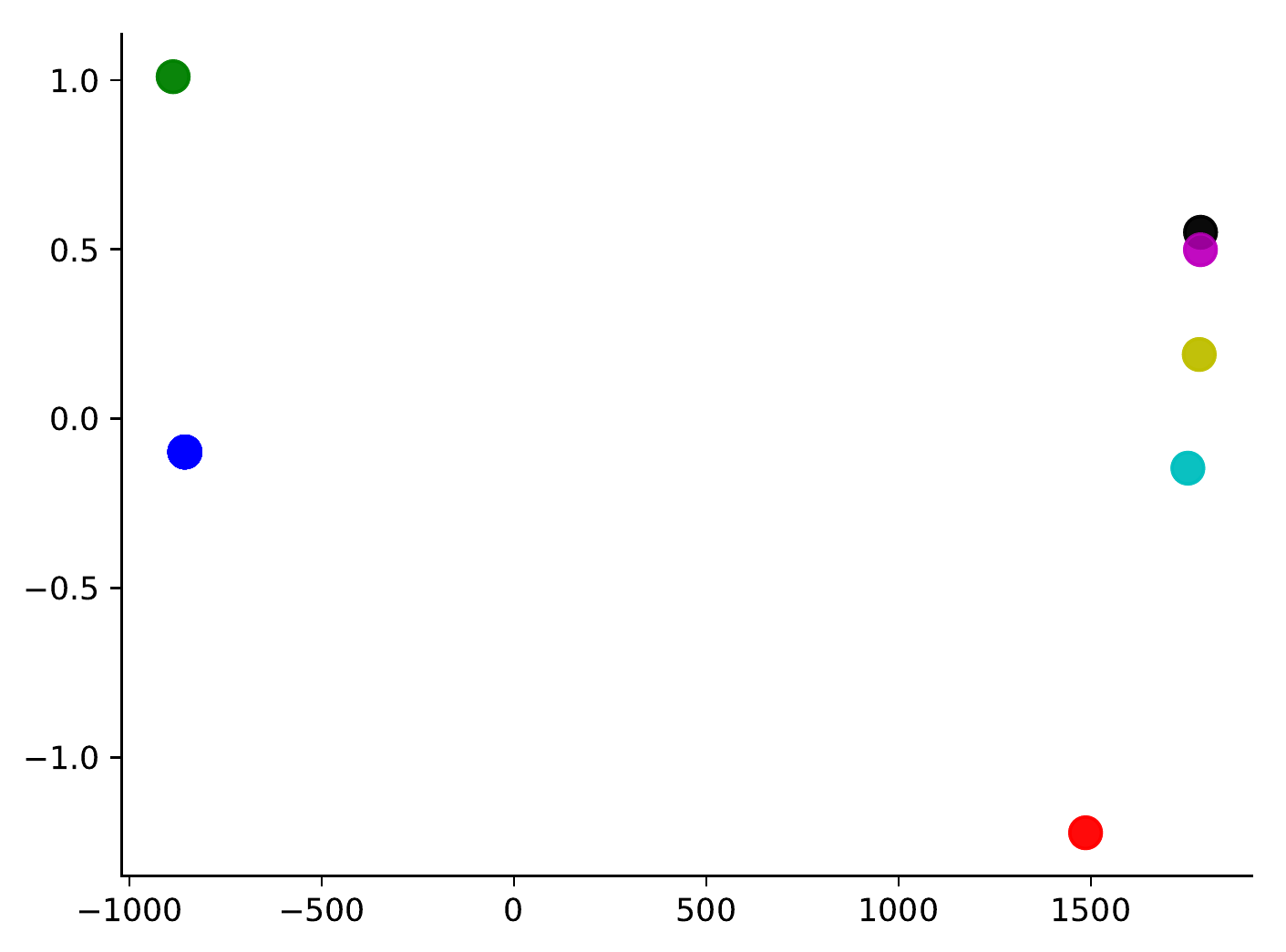}
  \end{minipage}
  \caption{The Barbell graph $B(10,10)$ (Top). Vertex representations in $\mathbb{R}^2$ learned by struc2vec (Bottom Left) and by the proposed kernel MPGK (Bottom Right).}
  \label{fig:barbell_embeddings}
\end{figure}

We use the proposed framework to embed the vertices of barbell graph in the $2$-dimensional space.
We denote as $B(h,k)$ the $(h,k)$-barbell graph which consists of two copies of the complete graph $K_h$ (each having $h$ vertices) that are joined by a path graph $P_k$ of length $k$.
Let $\{p_1,\ldots,p_k\}$ be the set of vertices of $P_k$.
Then, vertex $p_1$ is connected with an edge with one of the vertices of the one complete graph, while vertex $p_k$ is connected with an edge with one of the vertices of the second complete graph.
The $B(10, 10)$ graph is illustrated in Figure~\ref{fig:barbell_embeddings} (Top).
It is clear that many pairs of vertices of the $B(10, 10)$ graph have the same structural identity.
More specifically, vertices with the same color in the Figure are structurally equivalent.
For instance, all the vertices of the two complete graphs except from the two that are connected with $p_1$ and $p_k$ are structurally equivalent.
Permuting any pair of these vertices gives rise to an automorphism.

We expect the proposed framework to learn vertex representations that capture the structural equivalence illustrated in Figure~\ref{fig:barbell_embeddings} (Top).
Pairs of vertices that are structurally equivalent should be close to each other in the embeddings space.
Figure~\ref{fig:barbell_embeddings} shows the representations of the vertices of $B(10, 10)$ learned by struct2vec (Botton Left) and by MPGK RR (Bottom Right).
To learn these representations, we assigned an attribute to each vertex.
The attribute of a vertex was set equal to its degree.
In the first iteration, we used a linear kernel to compare two vertices (\ie the product of their degrees).
We then computed all the kernel values between vertices for $5$ more iterations and built the kernel matrix between the vertices.
Finally, we projected the vertices in the $2$-dimensional space using kernel PCA \cite{scholkopf1997kernel}.
Both the proposed framework and struct2vec managed to learn representations that properly separate the equivalent classes.
Specifically, the proposed kernel learned exactly the same representation for structurally equivalent vertices, while struc2vec placed such vertices close to each other in the embedding space.
Furthermore, the proposed kernel also captures structural hierarchies: the vertices belonging to the following two groups: ($1$) the two complete graphs, and ($2$) the path graph $P_k$, are very close to vertices belonging to the same group and very far from vertices belonging to the other group.

\subsection{Graph Classification}

\textbf{Datasets}.
We evaluated the proposed framework on standard graph classification datasets\footnote{The datasets, further references and statistics are available at \url{https://ls11-www.cs.tu-dortmund.de/staff/morris/graphkerneldatasets}} derived from bioinformatics and chemoinformatics (MUTAG, ENZYMES, NCI$1$, PROTEINS), and from social networks (IMDB-BINARY, IMDB-MULTI, REDDIT-BINARY, REDDIT-MULTI-$5$K, COLLAB).
We also demonstrated the effectiveness of the proposed framework on a synthetic dataset (Synthie).
Note that MUTAG and NCI$1$ contain graphs with discrete node labels, Synthie contains graphs with continuous node attributes, and ENZYMES and PROTEINS contain graphs with both discrete node labels and continuous node attributes.
On the other hand, the graphs contained in the social interaction datasets are unlabeled.

\textbf{Experimental Setup}.
To perform graph classification, we employed a C-Support Vector Machine (SVM) classifier.
We performed $10$-fold cross-validation, using $9$ folds for training and $1$ fold for testing.
The whole process was repeated $10$ times for each dataset and each kernel.
The parameter $C$ of the SVM was optimized on the training set only.

The parameters of the proposed message passing graph kernels were selected using cross-validation on the training dataset.
We chose the number of iterations $T$ from $\{ 1,2,3,4 \}$, which means that we computed $4$ different kernel matrices in each experiment.
We set parameters $\alpha$ and $\beta$ to $0.8$ and $0.2$ respectively.
Furthermore, we use the the Nystr{\"o}m method with $200$ samples to approximate the kernel matrix between vertices.
The proposed kernels were written in Python\footnote{Code available at \url{https://github.com/giannisnik/message_passing_graph_kernels}}.

We compare the proposed framework against several state-of-the-art kernels.
Specifically, our set of baselines include the GraphHopper kernel (GH) \cite{feragen2013scalable}, an instance of the graph invariant kernels (GI) \cite{orsini2015graph}, the propagation kernel (P2K) \cite{neumann2016propagation} and the hash Weisfeiler-Lehman subtree kernel (HGK-WL) \cite{morris2016faster}.
All these kernels support continuous vertex attributes.
Additionally, we compare the proposed message passing kernels to the Weisfeiler-Lehman subtree kernel (WL) \cite{shervashidze2011weisfeiler} and the shortest-path kernel (SP) \cite{borgwardt2005shortest}, which can only handle graphs with discrete node labels, to exemplify the usefulness of using continuous attributes.
For GH, GI, P2K and HGK-WL, we report the results from \cite{morris2016faster} since the experimental setup is the same with ours.
For WL and SP, we report the results from \cite{shervashidze2011weisfeiler}.
We also compare the proposed message passing kernels against two recent neural network architectures for graph classification: ($1$) PATCHY-SAN (PSCN $k=10$) \cite{niepert2016learning} and ($2$) Deep Graph Convolutional Neural Network (DGCNN) \cite{zhang2018end}.
For both neural network architectures, we report the best results from the corresponding papers since they were under the same setting as ours.

\textbf{Results}.
We report in Table~\ref{tab:classification_results} average prediction accuracies and standard deviations over the $10$ runs of the $10$-fold cross validation procedure.
\begin{table}[t]
\caption{Classification accuracy ($\pm$ standard deviation) of the proposed message passing graph kernels and the baselines on the $10$ graph classification datasets. NA indicates that results are not available.}
\label{tab:classification_results}
\centering
\def\arraystretch{1.1}
\scriptsize
\resizebox{\textwidth}{!} {
\begin{tabular}{|l|c|c|c|c|c|} \hline
\multirow{2}{*}{\backslashbox{Method}{Dataset}} & \multirow{2}{*}{MUTAG} & \multirow{2}{*}{ENZYMES} & \multirow{2}{*}{NCI$1$} & \multirow{2}{*}{PROTEINS} & \multirow{2}{*}{Synthie} \\
& & & & & \\ \hline
SP \cite{shervashidze2011weisfeiler} & 87.28 ($\pm$ 0.55) & 41.68 ($\pm$ 1.79) & 73.47 ($\pm$ 0.11) & NA & NA \\ 
WL \cite{shervashidze2011weisfeiler} & 82.05 ($\pm$ 0.36) & 52.22 ($\pm$ 1.26) & 82.19 ($\pm$ 0.18) & NA & NA \\ 
GH \cite{morris2016faster} & NA & 68.80 ($\pm$ 0.96) & NA & 72.26 ($\pm$ 0.34) & 73.18 ($\pm$ 0.77) \\ 
GI \cite{morris2016faster} & NA & \textbf{71.70} ($\pm$ 0.79) & NA & \textbf{76.88} ($\pm$ 0.47) & 95.75 ($\pm$ 0.50) \\ 
P2K \cite{morris2016faster} & NA & 69.22 ($\pm$ 0.34) & NA & 73.45 ($\pm$ 0.48) & 50.15 ($\pm$ 1.92) \\ 
HGK-WL \cite{morris2016faster} & NA & 67.63 ($\pm$ 0.95) & NA & 76.70 ($\pm$ 0.41) & 96.75 ($\pm$ 0.51) \\ 
PSCN $k=10$ \cite{niepert2016learning} & \textbf{88.95} ($\pm$ 4.37) & NA & 76.34 ($\pm$ 1.68) & NA & NA \\ 
DGCNN \cite{zhang2018end} & 85.83 ($\pm$ 1.66) & NA & 74.44 ($\pm$ 0.47) & NA & NA \\ \hline
MPGK RR & 85.26 ($\pm$ 1.16) & 44.38 ($\pm$ 1.36) & 60.12 ($\pm$ 0.17) & 60.03 ($\pm$ 0.12) & 77.80 ($\pm$ 0.94) \\ 
MPGK RA & 84.10 ($\pm$ 1.12) & 69.58 ($\pm$ 0.96) & 83.08 ($\pm$ 0.51) & 75.88 ($\pm$ 0.26) & 90.92 ($\pm$ 0.92) \\ 
MPGK AR & 84.80 ($\pm$ 2.33) & 48.58 ($\pm$ 1.51) & 71.50 ($\pm$ 0.37) & 72.91 ($\pm$ 0.35) & 89.35 ($\pm$ 1.12) \\ 
MPGK AA & 83.21 ($\pm$ 0.94) & 70.18 ($\pm$ 1.33) & \textbf{83.85} ($\pm$ 0.36) & 61.14 ($\pm$ 1.14) & \textbf{98.45} ($\pm$ 0.48) \\ \hline
\end{tabular}
}
\\
\vspace{0.1cm}
\resizebox{\textwidth}{!} {
\begin{tabular}{|l|c|c|c|c|c|} \hline
\multirow{2}{*}{\backslashbox{Method}{Dataset}} & IMDB & IMDB & REDDIT & REDDIT & \multirow{2}{*}{COLLAB} \\
& BINARY & MULTI & BINARY & MULTI-5K & \\ \hline
DGK \cite{yanardag2015deep} & 66.96 ($\pm$ 0.56) & 44.55 ($\pm$ 0.52) & 78.04 ($\pm$ 0.39) & 41.27 ($\pm$ 0.18) & 73.09 ($\pm$ 0.25) \\ 
PSCN $k=10$ \cite{niepert2016learning} & 71.00 ($\pm$ 2.29) & 45.23 ($\pm$ 2.84) & 86.30 ($\pm$ 1.58) & 49.10 ($\pm$ 0.70) & 72.60 ($\pm$ 2.15) \\ 
DGCNN \cite{zhang2018end} & 70.03 ($\pm$ 0.86) & 47.83 ($\pm$ 0.85) & NA & NA & 73.76 ($\pm$ 0.49) \\ \hline
MPGK RR & 67.15 ($\pm$ 0.94) & 29.50 ($\pm$ 0.49) & 74.67 ($\pm$ 0.39) & 47.64 ($\pm$ 0.10) & 65.52 ($\pm$ 0.14) \\ 
MPGK RA & 72.83 ($\pm$ 0.73) & 50.98 ($\pm$ 0.29) & \textbf{91.62} ($\pm$ 0.54) & \textbf{53.62} ($\pm$ 0.17) & 74.85 ($\pm$ 0.16) \\ 
MPGK AR & 72.64 ($\pm$ 0.51) & 49.40 ($\pm$ 0.38) & 84.57 ($\pm$ 0.40) & 52.87 ($\pm$ 0.43) & 68.95 ($\pm$ 0.28) \\ 
MPGK AA & \textbf{73.67} ($\pm$ 0.44) & \textbf{51.76} ($\pm$ 0.42) & 90.91 ($\pm$ 0.31) & 52.11 ($\pm$ 0.52) & \textbf{82.60} ($\pm$ 0.54) \\ \hline
\end{tabular}
}
\end{table}
The proposed message passing graph kernels outperform all baselines on $7$ out of the $10$ datasets.
The difference in performance between the proposed kernels and the baselines is larger on the social interaction datasets.
In some cases, the gains in accuracy over the best performing competitors are considerable.
For instance, on the REDDIT-BINARY, REDDIT-MULTI-$5$K, and COLLAB datasets, we offer respective absolute improvements of $5.32$, $4.52$, and $8.84$ in accuracy over the best competitor.
Furthermore, on almost all datasets, our message passing graph kernels reach better performance than the recent graph neural network architectures (PSCN and DGCNN), showing that kernels are still the dominant approach for the classification of small and medium-sized graph datasets.
It is interesting to note that the two kernels that operate on graphs with discrete node labels (SP and WL) fail to achieve performance comparable to kernels that use vertex attribute information (GH, GI, P2K, HGK-WL and proposed kernels) on the ENZYMES dataset (this dataset contains both discrete node labels and continuous node attributes).
This highlights the added advantage of kernels capable of handling continuous vertex attributes.
The proposed kernels outperform the baselines that use vertex attribute information on Synthie, and reach comparable performance on the two datasets that contain both discrete node labels and continuous node attributes (ENZYMES and PROTEINS).
It should be mentioned that the baseline kernels take into account both types of labels.
Our proposed framework is very general and we could have also designed variants that also take both types of information into account.
As regards the four variants of the proposed framework, MPGK AA was the best performing variant, while MPGK RA performed comparably on most datasets.
Both these kernels were generally superior than MPGK RR and MPGK AR.

\subsection{Molecular Graph Regression}

\textbf{Dataset}.
We evaluate the proposed framework on the publicly available QM9 dataset \cite{ramakrishnan2014quantum}.
The dataset contains approximately $134k$ organic molecules.
Each molecule consists of Hydrogen (H), Carbon (C), Oxygen (O), Nitrogen (N), and Flourine (F) atoms and contain up to $9$ heavy (non Hydrogen) atoms. 
Furthermore, each molecule has $13$ target properties to predict.
These properties can be grouped into $4$ categories: ($1$) those related to how tightly bound together the atoms in a molecule are (U0, U, H, G), ($2$) those related to fundamental vibrations of the molecule (Omega, ZPVE), ($3$) those that concern the states of the electrons in the molecule (HOMO, LUMO, gap), and ($4$) measures of the spatial distribution of electrons in the molecule (mu, alpha, R2).

\textbf{Experimental Setup}.
The dataset was divided into a train, a validation and a test set according to a $80\% / 10\% / 10\%$ split.
All target variables were normalized to have zero mean and unit variance.
Although the dataset contains spatial information related to the atomic configurations, in our experiments, we only used the graph representation of each molecule along with the attributes of the atoms (\ie vertex attributes).

To predict the targets, we only used MPGK AA, the kernel that performed best in graph classification.
We performed $4$ iterations in total.
We set parameters $\alpha$ and $\beta$ to $0.8$ and $0.2$ respectively.
Furthermore, we use the the Nystr{\"o}m method with $200$ samples to approximate the kernel matrix between vertices.
Due to the large size of the dataset, we did not compute the whole kernel matrix between graphs at each iteration, but we instead used the Nystr{\"o}m method with $200$ samples to approximate it.
Hence, we generated $4$ (one for each iteration) $200$-dimensional representations for each graph.
We concatenated these rerpesentations and fed them to a fully connected neural network with $512$ hidden units.
To train the model, we used the Adam optimizer with an initial learning rate of $0.001$.
The learning rate decayed linearly after each step towards a minimum of $10^{-6}$.
We set the number of epochs to $200$ after experimenting on the validation set. 
We trained the network separately for each target.

We compare the proposed message passing kernel against the  optimal assignment Weisfeiler–Lehman graph kernel \cite{kriege2016valid}, the convolutional neural network for learning molecular fingerprints (NGF) \cite{duvenaud2015convolutional}, PATCHY-SAN (PSCN $k=10$) \cite{niepert2016learning}, and the $2$nd order covariant compositional network ($2$nd order CCN) \cite{kondor2018covariant}.
For all baselines, we report the results from \cite{kondor2018covariant} since the experimental setup is the same with ours.

\textbf{Results}.
Table~\ref{tab:regression_results} illustrates the mean absolute error (MAE) and the the root mean squared error (RMSE) for the $13$ normalized targets.
\begin{table}[t]
\caption{Comparison of the baseline methods (Left) and the proposed message passing graph kernel (right) on the QM9 dataset.}
\label{tab:regression_results}
\centering
\def\arraystretch{1.1}
\scriptsize
\resizebox{\textwidth}{!} {
\begin{tabular}{|l|cc|cc|cc|cc|cc|} \hline
\multirow{3}{*}{\backslashbox{Target}{Method}} & \multicolumn{2}{|c|}{\multirow{2}{*}{WLGK}} & \multicolumn{2}{|c|}{\multirow{2}{*}{NGF}} & \multicolumn{2}{|c|}{PSCN} & \multicolumn{2}{|c|}{$2$nd order} & \multicolumn{2}{|c|}{\multirow{2}{*}{MPGK}} \\
& & & & & \multicolumn{2}{|c|}{$k=10$} & \multicolumn{2}{|c|}{CCN} & & \\ \cline{2-11}
& MAE & RMSE & MAE & RMSE & MAE & RMSE & MAE & RMSE & MAE & RMSE \\ \hline
mu & 0.69 & 0.92 & 0.63 & 0.87 & 0.54 & 0.75 & \textbf{0.48} & \textbf{0.67} & 0.51 & 0.79 \\
alpha & 0.46 & 0.68 & 0.43 & 0.65 & 0.20 & 0.31 & \textbf{0.16} & \textbf{0.26} & 0.20 & 0.31 \\ 
HOMO & 0.64 & 0.91 & 0.58 & 0.81 & 0.51 & 0.70 & \textbf{0.39} & 0.55 & \textbf{0.39} & \textbf{0.53} \\ 
LUMO & 0.70 & 0.84 & 0.65 & 0.79 & 0.59 & 0.73 & 0.53 & 0.68 & \textbf{0.22} & \textbf{0.32} \\ 
gap & 0.72 & 0.86 & 0.67 & 0.82 & 0.60 & 0.75 & 0.54 & 0.69 & \textbf{0.28} & \textbf{0.40} \\ 
R2 & 0.55 & 0.81 & 0.49 & 0.71 & 0.22 & 0.31 & \textbf{0.19} & \textbf{0.27} & 0.34 & 0.50 \\ 
ZPVE & 0.57 & 0.72 & 0.51 & 0.66 & 0.43 & 0.55 & 0.39 & 0.51 & \textbf{0.07} & \textbf{0.09} \\ 
U0 & 0.52 & 0.67 & 0.47 & 0.62 & 0.34 & 0.44 & 0.29 & 0.39 & \textbf{0.23} & \textbf{0.34} \\
U & 0.52 & 0.67 & 0.47 & 0.62 & 0.34 & 0.44 & 0.29 & 0.40 & \textbf{0.23} & \textbf{0.34} \\ 
H & 0.52 & 0.68 & 0.47 & 0.62 & 0.34 & 0.44 & 0.30 & 0.40 & \textbf{0.22} & \textbf{0.35} \\ 
G & 0.51 & 0.67 & 0.46 & 0.62 & 0.33 & 0.43 & 0.29 & 0.38 & \textbf{0.23} & \textbf{0.33} \\ 
Cv & 0.59 & 0.78 & 0.47 & 0.65 & 0.27 & 0.34 & 0.23 & 0.30 & \textbf{0.15} & \textbf{0.21} \\ 
Omega & 0.72 & 0.84 & 0.63 & 0.77 & 0.57 & 0.73 & 0.45 & 0.65 & \textbf{0.02} & \textbf{0.35} \\ \hline
\end{tabular}
}
\vspace{-.68cm}
\end{table}
MPGK achieves lower MAE and RMSE values than all the baselines on $10$ out of the $13$ targets, while it is outperformed by the $2$nd order CCN on the remaining $3$ targets.
In some cases, the improvement in performance is significant.
For example, the MAE of the $2$nd order CCN on the ZPVE target was $0.39$, while that of the proposed kernel was $0.07$.
Overall, the obtained results suggest that the proposed kernel is competitive with state-of-the-art methods.

\section{Conclusion}\label{sec:conclusion}
In this paper, we proposed a general framework for designing graph kernels.
The proposed kernel capitalizes on the well-known message passing scheme on graphs.
We derived four instances of the proposed framework, and showed through extensive experiments that these kernels are competitive with state-of-the-art methods.

\small

\bibliographystyle{plain}
\bibliography{biblio}

\end{document}